\documentclass[letterpaper, 10 pt, conference]{ieeeconf}  

\IEEEoverridecommandlockouts                              
\usepackage{color}
\usepackage{graphics}
\usepackage{graphicx}
\usepackage{url}
\usepackage{amsmath}
\usepackage{amsfonts}

\usepackage{mathabx} 
\usepackage{caption}
\usepackage{subcaption}
\usepackage{tabularx}

\usepackage{algorithm,algcompatible,amsmath}
\algnewcommand\INPUT{\item[\textbf{Input:}]}%
\algnewcommand\OUTPUT{\item[\textbf{Output:}]}%

\title{\LARGE \bf
Switching between Limit Cycles in a Model of Running Using Exponentially Stabilizing Discrete Control Lyapunov Function
}

\author{Pranav A. Bhounsule$^a$, Ali Zamani$^a$, Jason Pusey$^b$ 
\thanks{$^a$ Dept. of Mechanical Engineering, The University of Texas at San Antonio, 
       One UTSA Circle, San Antonio, TX 78249, USA. Corresponding author:
       {\tt\small pranav.bhounsule@utsa.edu}.  
       $^b$ U.S. Army Research Laboratory, Aberdeen Proving Grounds, Aberdeen, MD 21005, USA.
       This work was supported by an NSF grant IIS 1566463 to PAB.}%
}

\begin{document}

\maketitle
\thispagestyle{empty}
\pagestyle{empty}

\begin{abstract}
This paper considers the problem of switching between two periodic motions, also known as limit cycles, to create agile running motions. For each limit cycle, we use a control Lyapunov function to estimate the region of attraction at the apex of the flight phase. We switch controllers at the apex, only if the current state of the robot is within the region of attraction of the subsequent limit cycle. If the intersection between two limit cycles is the null set, then we construct additional limit cycles till we are able to achieve sufficient overlap of the region of attraction between sequential limit cycles. Additionally, we impose an exponential convergence condition on the control Lyapunov function that allows us to rapidly transition between limit cycles. Using the approach we demonstrate switching between 5 limit cycles in about 5 steps with the speed changing from 2 m/s to 5 m/s.
\end{abstract}

\section{Introduction}


The ability to quickly switch between periodic motions or limit cycles allows legged robots to demonstrate agility or the ability to quickly change velocity or direction \cite{duperret2016towards}.
Very little work has been done in synthesizing agile (non-periodic) gaits, even though a great amount of literature exists on the creation of periodic or steady state gaits. This work provides a technique for  constructing non-steady or agile gaits by sequentially composing steady-state gaits, thereby capitalizing on the extensive work done on steady state gaits. 

\section{Background and Related Work}
A straightforward technique to create agile gaits is to create individual controllers for periodic motion as well as for all possible transitions. For example, Santos and Matos \cite{santos2011gait} tuned nonlinear oscillators to generate different gaits and gait transitions in a quadrupedal robot.  Haynes and Rizzi \cite{haynes2006gaits} used a heuristic approach in which transitions were created by sequentially changing the controllers for pairs of leg from start to goal gait while maintaining static stability (center of mass projection is within the support polygon of the legs). Byl et al. \cite{byl2017mesh} pre-computed transition controllers between successive apex states and used these controllers to map out the reachable state space. The disadvantage of the method is that it requires additional transition controllers to switch between limit cycles.

\begin{figure}[tbp]
  \begin{center}
   \includegraphics[scale=0.75]{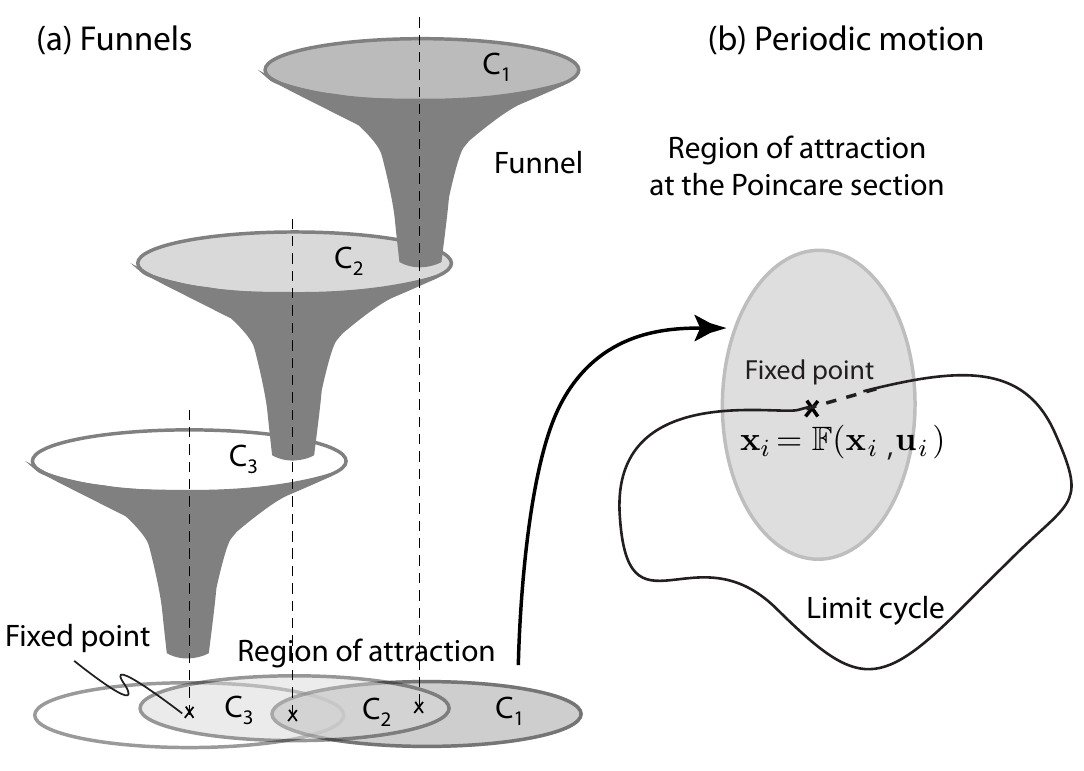}
 \end{center}
 \caption{Relation between funnels and limit cycles: (a) Use of funnels to sequentially compose motion \cite{burridge1999sequential}. (b) Running motion is analyzed using fixed point of the limit cycle at the Poincar\'e section. Switching controllers are created by composing limit cycles using the region of attraction to create funnels.}
\label{fig:funnels}
\end{figure}

\begin{figure*}[t]
  \begin{center}
   \includegraphics[scale=0.8]{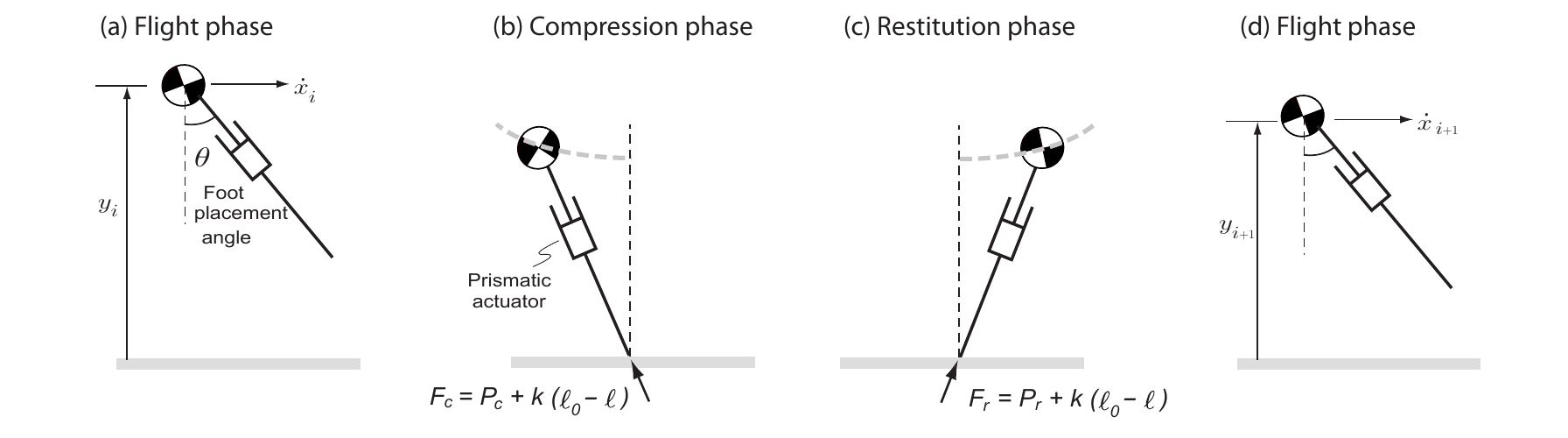}
 \end{center}
 \caption{Different phases of motion for the model. The model has a prismatic actuator and hip actuator (not shown) that can place the foot at an angle $\theta$ with respect to the vertical. The prismatic actuator can apply tensional force $F = P + k (\ell_0 - \ell)$ along the stance leg.   
 }
\label{fig:hopper}
\end{figure*}

Transition controller can be avoided entirely if one can find common states between two limit cycles and switch controllers when that particular state is reached. However, it is often very difficult to find common states between two limit cycles. As an alternative, it is relatively easier to use the region of attraction (range of states that converges to the limit cycle) \cite{strogatz1994nonlinear} 
to switch. We demonstrate the approach in Fig.~\ref{fig:funnels} (a). Each of the ellipsoids are the regions of attraction of a particular limit cycle. Initially we use the controller $C_1$ to get the system to move toward the fixed point corresponding to the first limit cycle. Once the state is inside the region of attraction of the next limit cycle, the controller $C_2$ is switched on and so on. This way, the system can funnel from one limit cycle to another \cite{burridge1999sequential}. 

Funnel-based switching has been used in the past and is also the main focus of this paper. 
Bhounsule et al. \cite{bhounsule2016dead} considered transition of a torso-actuated rimless wheel robot, a simple one degree of freedom system. The key idea was to use a one-step dead-beat control to switch from one fixed point to another in a single step. The switching was done at the mid-stance position using the measured velocity of the leg in contact with the ground. The region of attraction was not estimated but switching was attempted by trial and error. Westervelt \cite{westervelt2003switching} considered switching of a four degree of freedom, walking robot with knees and torso. To deal with the high dimensionality of the system, the switching was done by considering the zero dynamics of the system which is of dimension 1, simplifying the search for the region of attraction. Cao et al. \cite{cao2015control} considered the gait transition of a quadruped robot by switching in the mid-flight phase using the reduced state of the body position and velocity (that is, legs were excluded). The region of attraction was estimated by using forward simulation, which may be time consuming. Tedrake and coworkers \cite{tedrake2009lqr,tedrake2010lqr} used sum of squares to estimate the region of attraction and used Linear Quadratic Controllers to stabilize gaits, but the same idea can be extended to switch between limit cycles. Veer et. al. \cite{veer2017generation} used an exponential control Lyapunov function (ECLF) to switch between a continuum of limit cycles to generate variable speed walking. In their work, the ECLF ensures exponential local stabilization but in this work we use exponential orbital stabilization, which leads to a much faster transition.

In this paper, we use funnel-based approach for switching between limit cycles for a running model. The switching is done at the mid-stance position as shown in Fig~\ref{fig:funnels} (b). In contrast to past approaches, we use a discrete control Lyapunov function approach \cite{bhounsule2017,zamani2017} to: (1) estimate the region of attraction, and (2) enable exponential convergence to the fixed points. The latter allows for fast transitioning between limit cycles, typically in a single step, and is the main novelty of our approach.

\section{Model}
Figure~\ref{fig:hopper} shows the model of the runner that consists of a point mass body of mass $m$ 
and massless leg with a maximum leg length $\ell_0$.
Gravity points downwards and is denoted as $g$. 
There is a prismatic actuator that can generate an axial force ($F$) and a hip actuator for foot placement ($\theta$). Throughout this paper, we assume that the axial force is acting only when the leg contacts the ground and is given by the sum of a constant force term and a spring force term, that is, $F=P+k (\ell_0-\ell)$. 



The states of the model are given by $\{ x,\dot{x},y,\dot{y}\}$ where $x$ and $y$ are the x- and y-position of center of mass and $\dot{x}$ and $\dot{y}$ are the respective velocities. A single step of the walker is given below:
\begin{align}
\mbox{\scriptsize Flight} \underbrace{ \overbrace{\longrightarrow}^{\mbox{\scriptsize apex}} \mbox{\scriptsize Flight}  \overbrace{\longrightarrow}^{\mbox{\scriptsize touchdown}} \mbox{\scriptsize  Stance} \rightarrow \overbrace{\longrightarrow}^{\mbox{\scriptsize takeoff}} \mbox{\scriptsize Flight} }_{\mbox{\scriptsize one step/ period-one limit cycle}} \overbrace{\longrightarrow}^{\mbox{\scriptsize apex}} {\mbox{\scriptsize Flight}} \label{eqn:step}
\end{align}
The model starts at the apex where the state vector with respect to the world frame is, $\{0,\dot{x}_0,y_0,0\}$, a column vector.The model then falls under gravity, 
\begin{align}
\ddot{x} = 0, \mbox{            } \ddot{y} = -g, 
\end{align}
till contact with the ground is detected by the condition $y-\ell_0 \cos(\theta)=0$, where $\theta$ is the foot placement angle and measured relative to the vertical. Thereafter, the ground contact interaction is smooth and given by,
\begin{align}
 m \ddot{x} &= \bigg( P+k (\ell_0 - \ell) \bigg) \frac{x}{\ell},  \\
m \ddot{y} &= \bigg( P+k (\ell_0 - \ell) \bigg) \frac{y}{\ell} -m g,  \label{eqn:stance}
\end{align}
where $x$ and $y$ are taken relative to the contact point, $P>0$ is a constant thrust force, $k$ is the spring constant, and $\ell = \sqrt{x^2+y^2}$ is the instantaneous leg length. For the first half of the stance phase from touchdown to mid-stance (defined by $\dot{y} = 0$\footnote{Note that the event $\dot{y} = 0$ is different from the event corresponding to full leg compression, which is given by $\dot{\ell}=0$.}) called the compression phase, we assume that the constant force is $P = P_c$. For the second half of the stance phase from mid-stance to take-off called the restitution phase, we assume that the constant force is $P=P_r$. The takeoff takes place when the leg is fully extended, that is, $\ell_0 - \ell = 0$. Thereafter, the mass has a flight phase and ending up at the next apex state, $\{x,\dot{x},y,0\}$.

\section{Methods} \label{sec:methods}
Next, we describe our methodology for creating agile gaits. First, we create multiple limit cycles. Second, we define a Discrete Control Lyapunov Function (DCLF) for exponential decay of Lyapunov function between steps and estimate the region of attraction for each limit cycle. Third, we use the idea of funnels to transition from a start state to an end state by sequentially composing the limit cycles. 

\subsection{Poincar\'e map and limit cycle} \label{sec:poincare}
We define the Poincar\'e map, ${\bf \mathbb{F}}$, at the apex. The apex is defined by the condition, $\dot{y}=0$. Given the state at the apex at step $k$, ${\bf x}_k = \{\dot{x},y\}$, and the control, ${\bf u}_k = \{\theta,P_c,P_r\}$ (where $P_c$ and $P_r$ are constant forces during compression and restitution respectively (see Eqn.~\ref{eqn:stance})), we compute the state at the next step, 
\begin{align}
{\bf x}_{k+1} = {\bf  \mathbb{F}}({\bf x}_k,{\bf u}_k).
\end{align}
There is no closed form for the map $\mathbb{F}$. In this paper, it is found numerically by integrating the equations of motion. 
The $i$th limit cycle is found by fixing ${\bf x}_{k+1} = {\bf x}_k = {\bf x}_i$ and searching for ${\bf u}_k = {\bf u}_i = \{\theta,P_c = 0, P_r =0\}$ such that 
\begin{align}
{\bf x}_i = {\bf  \mathbb{F}}({\bf x}_i, {\bf u}_i).
\end{align}
The stability of the limit cycle can be found by evaluating the largest eigenvalue of the Jacobian of ${\bf  \mathbb{F}}$, that is, $J = \frac{\partial {\bf  \mathbb{F}}}{\partial {\bf x}} |_{({\bf x}_i,{\bf u}_i)}$ \cite{strogatz1994nonlinear}.  
For the runner one eigenvalue is always $1$ (a conservative system when $P_c = P_r = 0$) \cite{poulakakis2006stability} while the second eigenvalue is used to evaluate the stability. An eigenvalue less than $1$ indicates a stable limit cycle and unstable otherwise.


\subsection{Discrete Control Lyapunov Function (DCLF)}
We define a Lyapunov function for the $i$th limit cycle as follows 
\begin{align}
V(\Delta {\bf x}^i_k)=  (\Delta {\bf x}^i_k)^T {\bf S}_i \Delta {\bf x}^i_k = ({\bf x}_k-{\bf x}_i)^T {\bf S}_i ({\bf x}_k-{\bf x}_i)  \label{eqn:def_lyapunov}
\end{align}
where the positive definite matrix ${\bf S}_i = diag\{s_{i1},s_{i2}\}$. The resulting Lyapunov function is a 2-dimensional ellipse that has its major and minor axes along the horizontal velocity and height axes respectively. However, a more generic form that is symmetric and positive definite may also be used and it will have its major and minor axes at an angle to the coordinate axes. 

For the system to be asymptotically stable, the following condition needs to be satisfied 
\begin{align} 
V(\Delta {\bf x}^i_{k+1}) - V(\Delta {\bf x}^i_k) < 0
\end{align}

However, asymptotic convergence can be slow. To converge faster, we use an exponentially decaying condition on the Lyapunov function
\begin{align}
V(\Delta {\bf x}^i_{k+1} ) - V(\Delta {\bf x}^i_k ) &\leq -\alpha V(\Delta {\bf x}^i_k ), \label{eqn:DeltaV}  
\end{align}
where $0<\alpha<1$ is the rate of decay of the Lyapunov function between steps. Thus, the condition for exponential stability can be written as shown in Eqn.~\ref{eqn:finalDCLF}. 

\begin{figure*}[!htbp]
\normalsize
\setcounter{equation}{9} 
\begin{align}
& V(\Delta {\bf x}^i_{k+1} ) - (1 - \alpha) V(\Delta {\bf x}^i_k )  \leq 0
\hspace{0.5cm} \nonumber \\ 
\Longrightarrow
& ({\bf x}_{k+1}-{\bf x}_i)^T {\bf S}_i ({\bf x}_{k+1}-{\bf x}_i) - (1 - \alpha) ({\bf x}_k-{\bf x}_i)^T {\bf S}_i ({\bf x}_k-{\bf x}_i)  \leq 0, \nonumber \\
\Longrightarrow
& \bigg({\bf  \mathbb{F}}({\bf x}_k,{\bf u}_k)-{\bf x}_i\bigg)^T {\bf S}_i \bigg({\bf  \mathbb{F}}({\bf x}_k,{\bf u}_k)-{\bf x}_i\bigg) - (1 - \alpha) ({\bf x}_k-{\bf x}_i)^T {\bf S}_i ({\bf x}_k-{\bf x}_i)  \leq 0. 
\label{eqn:finalDCLF}
\end{align}
\setcounter{equation}{10}
\hrulefill
\vspace*{4pt}
\end{figure*}

\subsection{Region Of Attraction (ROA)}  \label{sec:ROA}
The Region Of Attraction (ROA), $\mathcal{R}$, of the controller is the set of all initial conditions ${\bf x}_k$ that would converge to the corresponding limit cycle, ${\bf x}_i$. In our case, we are interested in all ${\bf x}_k$ for which we can find ${\bf u}_k$ such that Eqn.~\ref{eqn:finalDCLF} is satisfied. To find the ROA, $\mathcal{R}_i$ for a given limit cycle we need to find level set, $({\bf x}_k-{\bf x}_i)^T {\bf S}_i ({\bf x}_k-{\bf x}_i) = c$ such that Eqn.~\ref{eqn:finalDCLF} is satisfied. We restrict ourselves to $c \leq 1$. 

First, we find ${\bf S}_i = diag\{s_{i1},s_{i2}\}$ such that level $({\bf x}_k-{\bf x}_i)^T  {\bf S}_i ({\bf x}_k-{\bf x}_i) = 1$ intersects the state constraint $y = \ell_0$ (a conservative estimate that prevents the leg from stubbing the ground at the Poincar\'e section, the flight phase apex position,  assuming the           leg is vertical and at its nominal length $\ell_0$).


Second, we find maximum value of $c$ such that $({\bf x}_k-{\bf x}_i)^T  {\bf S}_i ({\bf x}_k-{\bf x}_i) \leq c$ and exponential stabilization conditions given by Eqn.~\ref{eqn:finalDCLF} are satisfied. This is done numerically as follows: (1) fix $c$ to a small value and compute ${\bf x}_k$'s on the level set, (2) check if a ${\bf u}_k$ exists such that Eqn.~\ref{eqn:finalDCLF} holds using nonlinear optimization, (3) increase $c$ and repeat, (4) stop when at least one ${\bf x}_k$ is infeasible or when $c=1$. 

\subsection{Minimizing Energy Cost} \label{sec:cost}
When finding a controller for each state within the ROA, we minimize the Mechanical Cost Of Transport (MCOT) defined as energy used per unit weight per unit distance travelled
\begin{align}
 E_{\scriptsize{net}} &= E_{\theta} + E_{P_c} + E_{P_r} \nonumber \\ 
                        & = \int_{\tiny \mbox{step}} \bigg( |k (\ell_0 - \ell) \dot{\ell} | + |P_c \dot{\ell} | +  |P_r \dot{\ell} | \bigg) dt  \nonumber \\
\mbox{MCOT} & = \frac{E_{\scriptsize{net}}}{m g D}  \label{eqn:MCOT}
\end{align}
where $E_{\theta}$, $ E_{P_c}$, and $E_{P_r}$ are mechanical work done by spring force due to foot placement, constant compression force, and restitution force respectively, $|x|$ is the absolute value of $x$, $D$ is the horizontal distance travelled between two consecutive apex positions, and $\dot{\ell} = \frac{x \dot{x} + y \dot{y}}{\ell}$. The absolute value is a non-smooth function so we smooth it using square root smoothing \cite{srinivasan2006walk}. That is, $|x| = \sqrt{x^2+\epsilon^2}$ where $\epsilon$ is a small number (we assumed $\epsilon = 0.01$). The optimization problem is to minimize MCOT (Eqn.~\ref{eqn:MCOT}) subject to the exponential DCLF condition (Eqn.~\ref{eqn:finalDCLF}) for the given initial condition ${\bf x}_0$.

\subsection{Transitioning using funnels}
The key idea behind transitioning between limit cycles with stability guarantees is shown in Fig.~\ref{fig:funnels} and inspired by \cite{burridge1999sequential}. 
We choose limit cycles such that the fixed point of one limit cycle is in the region of attraction of the next limit cycle. When the system state is in the region of attraction of the next limit cycle, the corresponding controller is switched on. The algorithm is given in Algo.\ref{algorithm:transition}.

\begin{algorithm}
    \caption{Transition(initial state ${\bf x}_{\scriptsize init}$, goal state ${\bf x}_{\scriptsize goal}$)} 
  \begin{algorithmic}[1]
    \INPUT initial state  ${\bf x}_{\scriptsize init}$, goal state ${\bf x}_{\scriptsize goal}$. 
    \OUTPUT Set of control actions for each step, $u_k$, where step index is $k =1,2,...,n$.
    \STATE Compute sufficiently large number of limit cycles (say $m$) and their associated controller 
    ($u$) between init and goal state such that their ROA's ($\mathcal{R}$) overlap (See Secs.~\ref{sec:poincare} and \ref{sec:ROA}).
    \STATE SORT($\mathcal{R}_p$), $p=1,2,...,m$ either in increasing or decreasing order of speed of fixed point.
    \STATE  Initialize step number $ k \leftarrow 0$ and state ${\bf x}_{0}  \leftarrow {\bf x}_{\scriptsize init}$ 
       \WHILE{ $ |x_{k+1} - {\bf x}_{\scriptsize goal} |  \geq \delta$} 
       \COMMENT {$\delta$ is a small number}
     \STATE FIND(${\bf x_j}$) all the fixed points ${\bf x}_j$ where $j=1,2,...,N$  such that ${\bf x}_{k} \in \mathcal{R}_j$ using Eqn.~\ref{eqn:def_lyapunov}.  \COMMENT{NOTE: ${\bf x}_j$ is a subset of the $m$ limit cycles computed earlier.}
     \STATE Choose fixed point ${\bf x}_j$ that is closest to the goal state (say ${\bf \bar{x}}_j$)  and set the corresponding controller $u_k= {\bf {\bar u}}_j$.
     \STATE Apply the control $u_k$ and compute the state at the next apex, ${\bf x}_{k+1}$.
     \STATE $k \leftarrow k+1$
  \ENDWHILE
  \end{algorithmic}
  \label{algorithm:transition}
\end{algorithm}

\section{Results} \label{sec:results}
In all the following results we assume that mass is $m=80$ $kg$, nominal leg length is $\ell_0=1$ $m$, gravity $g=10$ $m/s^2$, and spring constant $k=32000$ $N/m$. For all simulations, we choose the exponential decay rate $\alpha=0.9$ (see Eqn.~\ref{eqn:finalDCLF}).

\subsection{Example 1: Exponential Stabilization}
To illustrate the exponential stabilization of the controller we proceed as follows. 
First, we found foot placement $\theta$ needed to achieve a fixed point of ${\bf x}^* = \{ 5.0,1.3\}$. The corresponding value is $\theta^* = 0.3465$ rad. The maximum eigenvalue was $1.33$ indicating an unstable gait. We found the positive definite matrix for the Lyapunov function to be ${\bf S}^* = diag\{1, 11.1\}$ as described in Sec.~\ref{sec:ROA}. To test the controller, we started with an initial condition at step 0, ${\bf x}_0 = \{ 4.20, 1.48 \}$, that is different from the fixed point. The value of the Lyapunov function for the initial condition is $V(\Delta {\bf x}_0) = (\Delta {\bf x}^*_0)^T {\bf S}^*_i \Delta {\bf x}^*_0 = 1$. Subsequently, the controller ensures exponential decay of the Lyapunov function, $V(\Delta {\bf x}_1) = 0.1$ and $V(\Delta {\bf x}_2) = 0.01$. 
%
%
%
%
%
%
%
%
%
%
%
%
%
%
%
%
%
%
%
%

%

\subsection{Example 2: Robustness to height variation}
\begin{table*}[!htbp]
\caption{Transition for example 2: Robustness to height change. The fixed point of the target limit cycle is $\{ 2, 1.3\}$. The nominal foot placement angle is $0.1603$ rad. A ditch of height $0.2$ m is introduced due to which the system starts with the initial condition, ${\bf x}_0 = \{1.957,1.5085\}$. The mechanical work done by the actuator due to spring force for the limit cycle is $E^i_\theta=795.4332$ $J$ and $\mbox{MCOT}^i= 0.7533$.}
\label{tab:limit_cycle3}
\begin{center}
\begin{tabular}{| c | c | c | c | c | c | c | c | c | c |}
\hline
$k$ & $i$ & ${\bf x}_k$ & $\theta$ & $P_c$ &  $P_r$ & $E_\theta-E^i_\theta$ & $E_{P_c}$ & $E_{P_r}$ & MCOT\\
\hline
$0$ & $ 1$ & $\{1.957,1.5085\}$ & $0.14823$ & $941.2727$ & $0$ & $54.4924$ & $153.3961$ & $0$ & $ 0.89718$\\
$1$ & $ 1$ & $\{1.8279,1.3412\}$ & $0.13238$ & $0$ & $0$ & $49.7985$ & $0$ & $0$ & $ 0.81926$\\
$2$ & $ 1$ & $\{1.9578,1.3166\}$ & & & & & & & \\
\hline
\end{tabular}
\end{center}
\end{table*}

\begin{figure}[tbp]
  \begin{center}
   \includegraphics[scale=0.4]{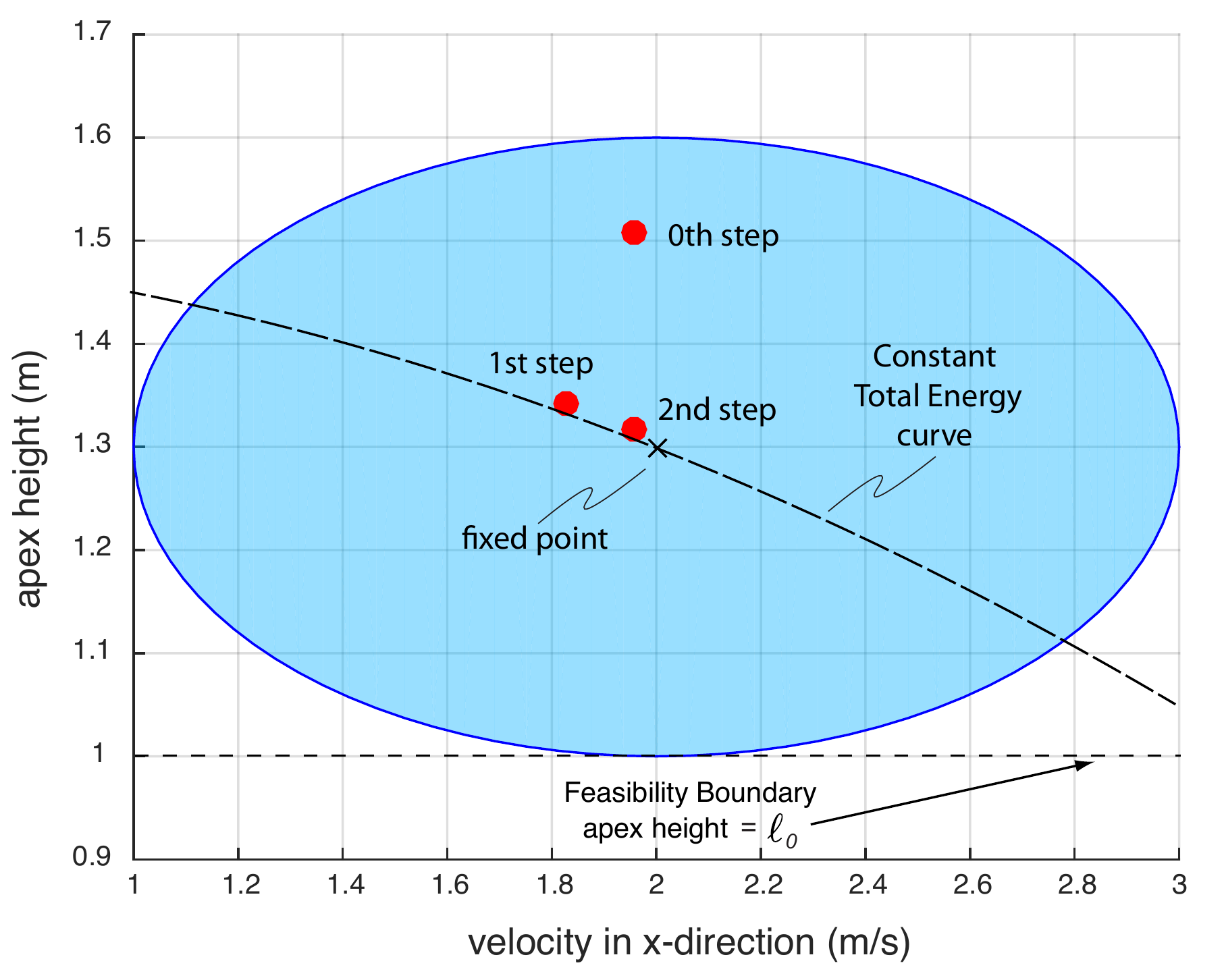}
 \end{center}
 \caption{Example 2, robustness to step down. A step down leads to increase in apex height because the height is measured relative to the ground. The black cross indicates the fixed point. The dashed line indicates the constant total energy, sum of kinetic and potential energies, that passes through the fixed point. The blue region indicates the ROA. The red dots indicate the system state at each step. The figure demonstrates how DCLF is able to tackle an external disturbance.}
\label{fig:phase_multiple3}
\end{figure}

Figure~\ref{fig:phase_multiple3} and Tab.~\ref{tab:limit_cycle3} demonstrate robustness of the DCLF. First, we found foot placement $\theta$ needed to achieve a fixed point of ${\bf x}^* =\{\dot{x}^*,y^*\}=\{ 2.0,1.3\}$. The corresponding value is $\theta^* = 0.1603$. To test the controller, we introduced a ditch of height $0.2$ $m$. Subsequently, the state of the system at the apex (step 0) is ${\bf x}_0 = \{1.957,1.5085\}$. 
Fig.~\ref{fig:phase_multiple3} shows the evolution of the state at subsequent steps. The black dashed line indicates the constant total energy line ($\mbox{TE}^* = mg y^* + 0.5 m \dot{x}^{*2}$) passing through the fixed point (black cross). Initially at step $0$ the $\mbox{TE}$ is greater than $\mbox{TE}^*$, thus energy needs to be dissipated. DCLF achieves this by using a non-zero force in compression phase ($P_c$) (see Tab.~\ref{tab:limit_cycle3}, row 1). Thereafter, on step $1$, the $\mbox{TE}$ reaches the constant TE line passing through the fixed point as shown in Fig.~\ref{fig:phase_multiple3}. The controller then adjusts only foot placement to further reduce the Lyapunov function (see Tab.~\ref{tab:limit_cycle3}, row 1).

\subsection{Example 3: Transitioning between limit cycles with and without actuator bounds}
\begin{figure*}[!htbp]
   \includegraphics[scale=0.45]{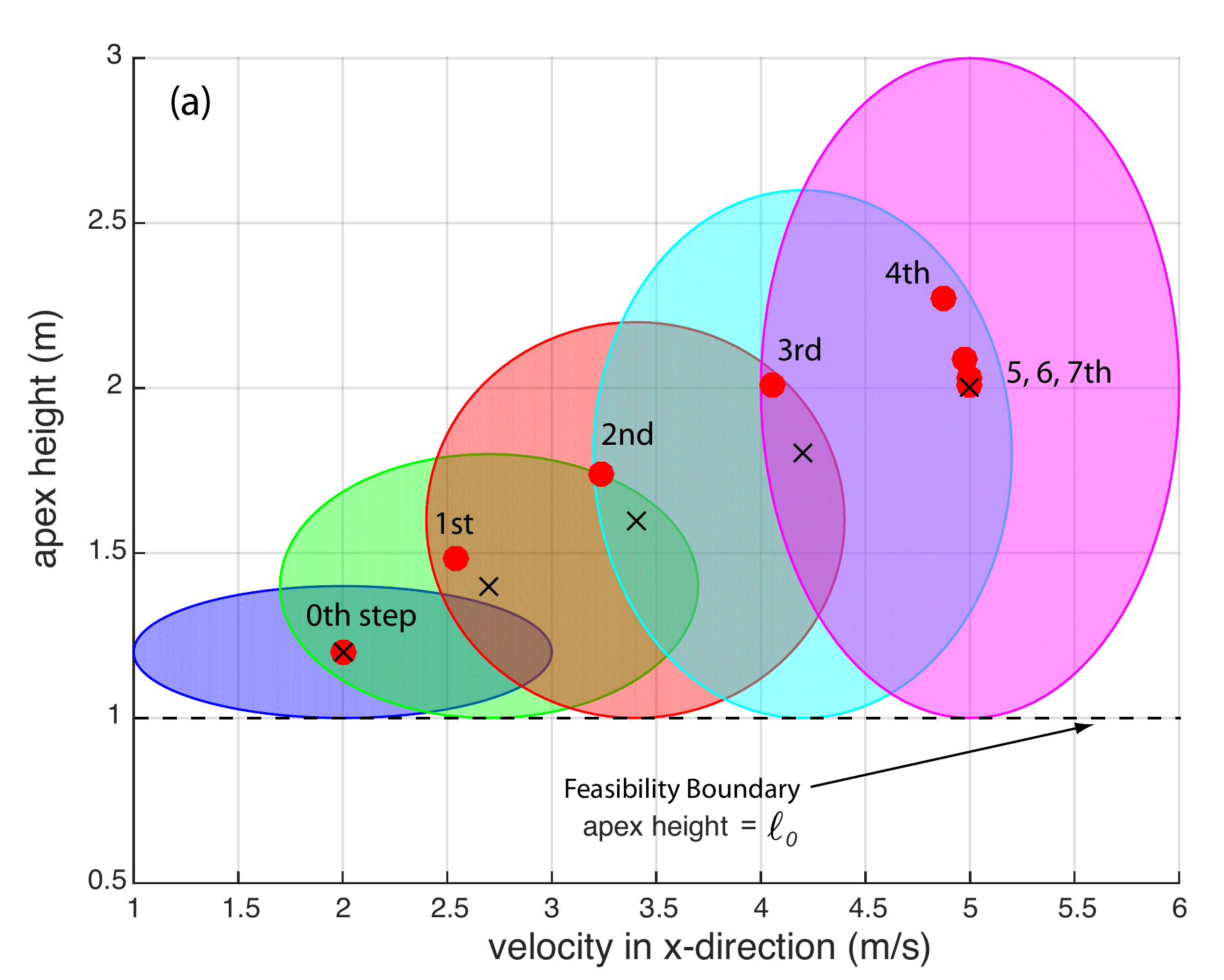} \includegraphics[scale=0.45]{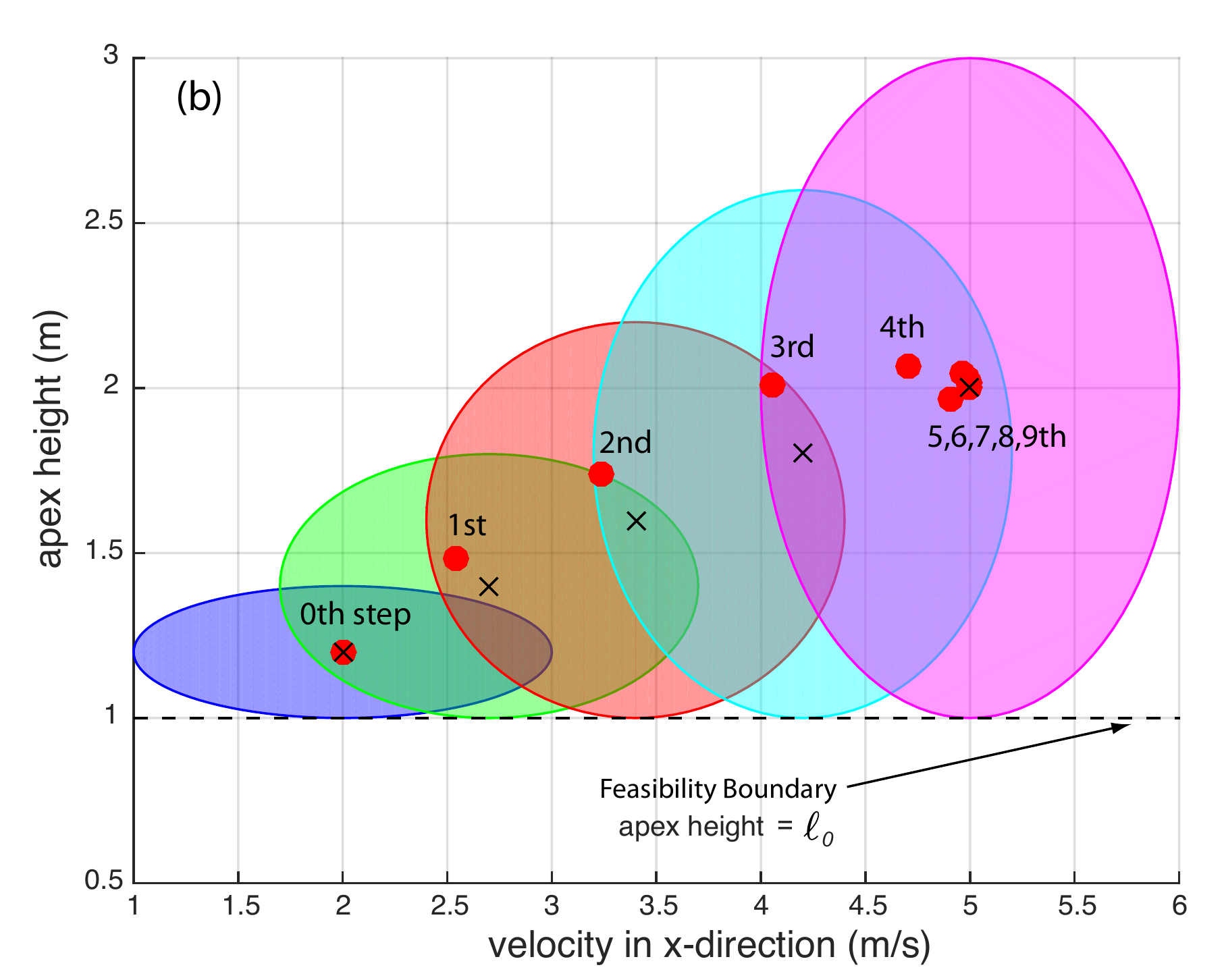}
 \caption{Transitioning between limit cycles for example 3 for (a) no actuator limits, (b) with actuator limits.  The black cross indicates the fixed point. The shaded region indicates the ellipsoid estimating the ROA for each limit cycle. The red dots indicate the system state at each step. An animation is in the reference \cite{bhounsule2017agility}
 }
\label{fig:phase_multiple1}
\end{figure*}
Figure~\ref{fig:phase_multiple1} illustrates how to switch between limit cycles using funnels. The model starts at the fixed point $\{2,1.2\}$ and needs to end at the fixed point $\{5,2\}$. We create limit cycles for these two fixed points and estimate their ROA. We also create additional limit cycles and estimate their ROA. We need a total of $3$ more limit cycles such that fixed point of one limit cycle is in the ROA of the subsequent limit cycle so that transitions are possible. Table~\ref{tab:limit_cycle1} shows the $5$ limit cycles, the foot placement angle, associated energy, and mechanical cost of transport.

We considered two cases: (a) no actuator bounds and (b) with actuator bound of $12 mg$ in the prismatic actuator. Figure~\ref{fig:phase_multiple1} shows the evolution of the state at the apex between successive steps while Tab.~\ref{tab:limit_cycle1b} shows in addition, the control strategy and energy usage. We can see that for steps $0$-$2$ the state, control, and energy are identical (compare rows 1, 2, and 3 in Tab.~\ref{tab:limit_cycle1b} (a) with those in (b)). Thereafter, the actuator limit kicks in and the convergence for (b) (with actuator limits) is slower than (a) (no actuator limits). Overall, it takes only $7$ steps for (a) but $9$ steps for (b) to be sufficiently close to the fixed point of the target limit cycle.  It is interesting to note that for both cases only the restitution force is applied (i.e., $P_r > 0$) in the first 4 steps to enable fast transition between limit cycles, as this has the effect of adding energy and speeding up the runner. 
The MCOT does not change appreciably in the entire transition.




\begin{table*}[htbp]
\caption{Limit cycles for example 3, transition controllers}
\label{tab:limit_cycle1}
\begin{center}
\begin{tabular}{| c | c | c | c | c | c | c | c |}
\hline
limit cycle, \# $i$  & ${\bf x}^i$ & $\mbox{maximum eigenvalue}$ & $\theta$ & $\{s_{i1},s_{i2}\}$ & $E^i_{\theta}$ & \mbox{MCOT} \\
\hline
$1$ & $\{2 , 1.2\}$  & $1.5958$ & $0.16328$ & $\{ 1 , 25 \}$ &  $588.9249$ & $0.6394$\\
$2$ & $\{2.7 , 1.4\}$  & $1.7246$ & $0.20813$ & $\{ 1 , 6.25 \}$ &  $1040.5214$ & $0.6564$\\
$3$ & $\{3.4 , 1.6\}$  & $1.8223$ & $0.25237$ & $\{ 1 , 2.7778 \}$ &  $1504.095$ & $0.6446$\\
$4$ & $\{4.2 , 1.8\}$  & $1.8846$ & $0.30156$ & $\{ 1 , 1.5625 \}$ &  $2003.4739$ & $0.6187$\\
$5$ & $\{5 , 2\}$  & $1.9269$ & $0.34897$ & $\{ 1 , 1 \}$ &  $2533.3107$ & $0.5987$\\
\hline
\end{tabular}
\end{center}
\end{table*}

\begin{table*}[htbp]
\caption{Transition for example 3: The source state is $\{2,1.2\}$ and target state is $\{ 5, 2\}$.}
\subcaption*{(a) No actuator limits}
\label{tab:limit_cycle1b}
\begin{tabular}{| c | c | c | c | c | c | c | c | c | c |}
\hline
step \#, $k$ & limit cycle, \# $i$  & ${\bf x}_k$ & $\theta$ & $P_c$ &  $P_r$ & $E_\theta-E^i_\theta$ & $E_{P_c}$ & $E_{P_r}$ & MCOT\\
\hline
$0$ & $ 2$ & $\{2,1.2\}$ & $0.10619$ & $0$ & $2561.2099$ & $-516.3278$ & $0$ & $327.0972$ & $ 0.6985$\\
$1$ & $ 3$ & $\{2.5369,1.4871\}$ & $0.14125$ & $0$ & $1960.9835$ & $-409.3977$ & $0$ & $362.1299$ & $ 0.73876$\\
$2$ & $ 4$ & $\{3.239,1.737\}$ & $0.19017$ & $0$ & $2033.0225$ & $-385.6915$ & $0$ & $456.4176$ & $ 0.71919$\\
$3$ & $ 5$ & $\{4.0495,2.0121\}$ & $0.24826$ & $0$ & $1895.4544$ & $-279.4241$ & $0$ & $502.4145$ & $ 0.69664$\\
$4$ & $ 5$ & $\{4.8733,2.2726\}$ & $0.32461$ & $363.5763$ & $0$ & $325.6946$ & $108.6778$ & $0$ & $ 0.66728$\\
$5$ & $ 5$ & $\{4.9688,2.0898\}$ & $0.34128$ & $137.2408$ & $0$ & $176.4281$ & $39.9361$ & $0$ & $ 0.63788$\\
$6$ & $ 5$ & $\{4.9912,2.0287\}$ & $0.34654$ & $45.8814$ & $0$ & $129.3415$ & $13.2341$ & $0$ & $ 0.62852$\\
\hline
\end{tabular}
\subcaption*{(b) With actuator limits.}
\label{limit_cycle1b}
\begin{center}
\begin{tabular}{| c | c | c | c | c | c | c | c | c | c |}
\hline
step \#, $k$ & limit cycle, \# $i$ & ${\bf x}_k$ & $\theta$ & $P_c$ &  $P_r$ & $E_\theta-E^i_\theta$ & $E_{P_c}$ & $E_{P_r}$ & MCOT\\ 
\hline
$0$ & $ 2$ & $\{2,1.2\}$ & $0.10619$ & $0$ & $2561.21$ & $-516.3278$ & $0$ & $327.0972$ & $ 0.6985$\\
$1$ & $ 3$ & $\{2.5369,1.4871\}$ & $0.14125$ & $0$ & $1960.9766$ & $-409.3978$ & $0$ & $362.1286$ & $ 0.73876$\\
$2$ & $ 4$ & $\{3.239,1.737\}$ & $0.19017$ & $0$ & $2033.0298$ & $-385.6945$ & $0$ & $456.4188$ & $ 0.71919$\\
$3$ & $ 5$ & $\{4.0495,2.0121\}$ & $0.25399$ & $0$ & $1017.2232$ & $-252.895$ & $0$ & $271.3035$ & $ 0.68289$\\
$4$ & $ 5$ & $\{4.7062,2.0638\}$ & $0.31812$ & $0$ & $0$ & $91.0704$ & $0$ & $0$ & $ 0.64186$\\
$5$ & $ 5$ & $\{4.9116,1.965\}$ & $0.34044$ & $0$ & $286.4697$ & $-3.8926$ & $0$ & $80.539$ & $ 0.62526$\\
$6$ & $ 5$ & $\{4.9883,2.0277\}$ & $0.34996$ & $0$ & $0$ & $171.1558$ & $0$ & $0$ & $ 0.63364$\\
$7$ & $ 5$ & $\{4.9607,2.0414\}$ & $0.34373$ & $19.6883$ & $0$ & $154.9589$ & $5.7058$ & $0$ & $ 0.63215$\\
$8$ & $ 5$ & $\{4.9949,2.0173\}$ & $0.34752$ & $27.9249$ & $0$ & $120.6562$ & $8.0413$ & $0$ & $ 0.62679$\\
%
\hline
\end{tabular}
\end{center}
\end{table*}

\section{Discussion}
We have presented a method for switching between limit cycles based on two key ideas: (1) use the region of attraction at the Poincar\'e section to create funnels, and (2) use exponentially stabilizing discrete control Lyapunov function for fast switching between limit cycles. We demonstrate the approach on a model of running with an axial actuator to control the ground interaction forces and a hip actuator to control the foot placement position. We show that a fairly wide range of initial perturbations can converge to within 1\% of the fixed point in a maximum of 2 steps. We also demonstrate the runner can transition from the limit cycle with a speed of $2$ $m/s$ to $5$ $m/s$ (a change of $150$$\%$) in about $5$ steps even with actuator limits. 

The use of three control actions ($P_c$, $P_r$, $\theta$) substantially increases the region of attraction of the model (e.g., the ellipsoid shown in Fig.~\ref{fig:phase_multiple3}). When the two constant forces are assumed to be zero (i.e., $P_c = P_r=0$) the region of attraction shrinks to a curve (e.g., the total energy dashed line shown in Fig.~\ref{fig:phase_multiple3}). This special case in which energy is conserved between steps is called the Spring Loaded Inverted Pendulum (SLIP) model of running \cite{schwind1998spring}.

The use of a discrete control Lyapunov function (DCLF) with exponential stabilization accelerates the convergence to the fixed point, thus enabling fast switching between limit cycles. This is a distinct advantage over controllers that impart asymptotic convergence, which is slower \cite{grizzle2001asymptotically}. Another approach is to use a one-step dead-beat control (full correction of disturbance in a single step) as it gives the fastest switching between limit cycles. However, we have found that DCLF can handle a larger range of modeling errors than a one-step dead-beat control and is thus preferred, especially when implementing on hardware \cite{bhounsule2017}.


Finally, we discuss limitations of our work. To enable transition, we need to have sufficient overlap between the region of attraction of sequential limit cycles. However, if the region of attraction is small then one would need to use a large number of limit cycles to transition thus increasing the computational complexity. 
The use of the nonlinear optimization to compute transition controllers is computationally expensive and online optimization might be challenging in practice especially for higher degrees of freedom robots. Alternately, the controllers can be saved as a look up table. 

\section{Conclusions}
In this paper we have demonstrated that steady state gaits can be sequentially composed to create non-steady or agile gaits. This is achieved by funneling the robot from the start state to goal state by using the region of attraction of successive, overlapping steady state limit cycles. Furthermore, a discrete control Lyapunov function with exponential decay is an effective method of enabling quick transition between limit cycles.






%

%

\bibliographystyle{IEEEtran}
\bibliography{pranav_bib2} 

%
%
%
%

\end{document}